\documentclass{article}

     \PassOptionsToPackage{numbers, compress}{natbib}

     \usepackage[preprint]{neurips_2020}

\usepackage[utf8]{inputenc} 
\usepackage[T1]{fontenc}    
\usepackage[hidelinks]{hyperref}       
\usepackage{url}            
\usepackage{booktabs}       
\usepackage{amsfonts}       
\usepackage{nicefrac}       
\usepackage{microtype}      
\usepackage{natbib}

\usepackage{inconsolata}
\usepackage{times}
\usepackage{latexsym}
\usepackage{graphicx}
\usepackage{latexsym}
\usepackage{amsmath}
\usepackage{amsthm}
\usepackage{booktabs}
\usepackage{algorithm}
\usepackage{algorithmic}
\usepackage[switch]{lineno}
\usepackage{multirow}
\usepackage{paralist}
\usepackage{tcolorbox}

\title{\textsc{InfFeed}: Influence Functions as a Feedback to Improve the Performance of Subjective Tasks}

\author{Somnath Banerjee,
  Maulindu Sarkar, Punyajoy Saha, Binny Mathew, Animesh Mukherjee \\
  Indian Institute of Technology Kharagpur, India \\
  \texttt{som.iitkgpcse@kgpian.iitkgp.ac.in} \\
  }

\begin{document}

\maketitle

\begin{abstract}
Recently, \textit{influence functions} present an apparatus for achieving explainability for deep neural models by quantifying the perturbation of individual train instances that might impact a test prediction. Our objectives in this paper are twofold. First we incorporate influence functions as a feedback into the model to improve its performance. Second, in a dataset extension exercise, using influence functions to automatically identify data points that have been initially `silver' annotated by some existing method and need to be cross-checked (and corrected) by annotators to improve the model performance.   
To meet these objectives, in this paper, we introduce \textsc{InfFeed}, which uses influence functions to compute the influential instances for a target instance. Toward the first objective, we adjust the label of the target instance based on its influencer(s) label. In doing this, \textsc{InfFeed} outperforms the state-of-the-art baselines (including LLMs) by a maximum macro F1-score margin of almost $4$\% for hate speech classification, $3.5$\% for stance classification, and $3$\% for irony and $2\%$ for sarcasm detection. Toward the second objective we show that manually re-annotating only those silver annotated data points in the extension set that have a negative influence can immensely improve the model performance bringing it very close to the scenario where all the data points in the extension set have gold labels. This allows for huge reduction of the number of data points that need to be manually annotated since out of the silver annotated extension dataset, the influence function scheme picks up $\sim\frac{1}{1000}$ points that need manual correction.
\end{abstract}

\section{Introduction}

In most of the classification problems, the real-world data (training and test instances) are not evenly distributed into classes~\cite{Bengioetal19}. As a result, the performance of the model suffers significantly, providing motivation to use pre-trained large-scale models. Despite these large models' excellent performance, most deep neural architectures are implemented as a black box and lack algorithmic transparency~\cite{Transparency}. Transparency in the method improves the explainability of the model and makes it more trustworthy.
\begin{figure}[t]
\centering
\includegraphics[width=0.5\textwidth]{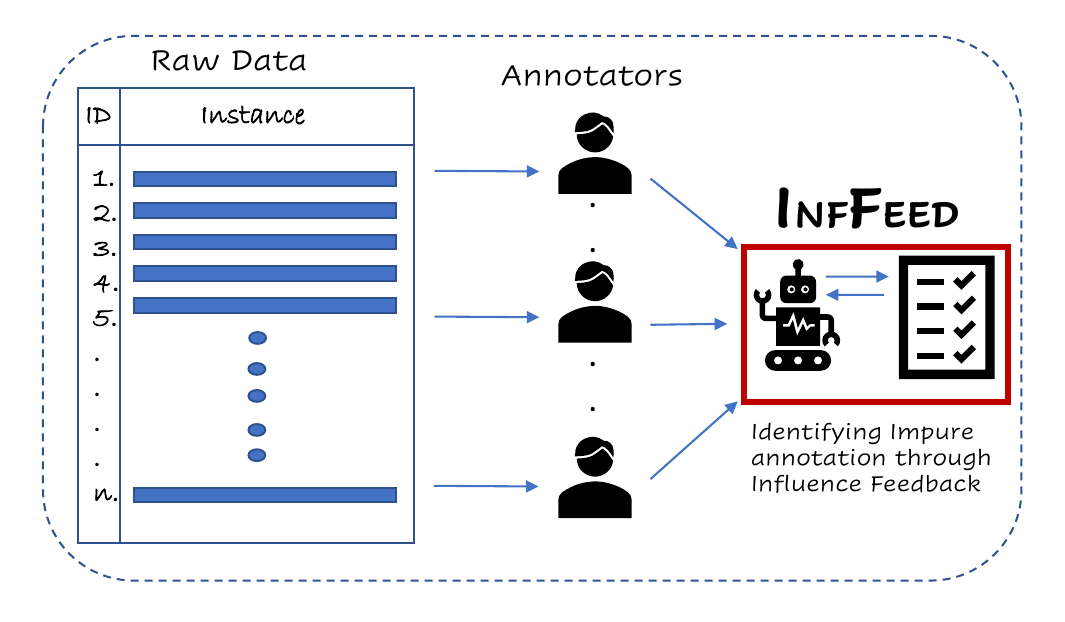}
\caption{Schematic illustrating our idea of using influence functions to revise the annotations of the target instance.}
\label{fig1}
\end{figure}
Some previous works attempt to explain the predictions of a model (i.e., why the model takes a particular decision) by perturbing the train instances or locally fitting the model on train data~\cite{Ribeiro:2016}. In addition, to explain the model, the authors in~\cite{pmlr-v70-koh17a} formulate influence functions to understand how the model predictions are affected by up-weighting a small amount of training instance loss. The idea is to estimate how much each training sample affects the model's predictions over the test set. Any training sample that causes the test loss to go up is considered less useful and is down-weighted afterward. Given the efficacy of influence-based data resampling in this work, we set a twofold objective. First we show that influence functions can be passed as a feedback to the model to improve its overall performance. Second, for the purposes of extension of annotated datasets, we show that influence functions can automatically identify those data points whose labels need to be cross-checked (and corrected) by annotators out of the full extension set that have been initially `silver' annotated by some existing model.

 Our main contributions to this paper are as follows.
\begin{compactitem}
    \item  We propose a framework called \textsc{InfFeed} where we employ the influence function as feedback to adjust the label of a candidate data point based on the labels of its influencers in order to increase the performance of the model. 
\item We evaluate the proposed framework on six datasets which are on subjective tasks such as hate speech detection, stance classification, irony, and sarcasm detection. 
\item We observe that our framework results in an improvement of 4\%, 3.5\%, 3\% and 2\% F1 score in the model performance over state-of-the-art baselines for hate speech, stance and irony, and sarcasm classification, respectively. 
\item For the dataset extension exercise, we show that just manually correcting the labels of the data points that impart a negative influence can result in a performance very close to the case where the whole extension set is gold annotated. The reduction is huge since the negatively influencing set is $\sim \frac{1}{1000}^\textrm{th}$, of the size of the full extension set. 
\end{compactitem}
This, we believe, is a first-of-its-kind approach to use influence functions play the role of a pseudo-annotator deciding whether to update the label of target instances in a \textit{text classification} model in order to improve its performance over state-of-the-art baselines. 

\section{Related work}

One of the most critical issues with deep learning models is their interpretability~\cite{Guidotti:2018,Lipton:2018}, and the proneness to learn ambiguous correlations instead of understanding the true nature of the task~\cite{Sagawa:2020}. These two reasons result in poor outcomes on datasets and cannot meet the expectations~\cite{Gururangan:2018,Jia:2017,Glockner:2018} resulting in severe biases in model decisions~\cite{Blodgett:2020,Sun:2019}. This further brings down the overall confidence in the technology~\cite{Ribeiro:2016,Ehsan:2019}. Despite great success, the question of ``why does the model predict what it predicts?'' needs a succinct answer. A satisfactory answer to this question can result in the improvement of the model~\cite{Amershi:2015}, lead to the development of newer perspectives~\cite{Srikumar:2017}, and benefit users by providing explanations of the model actions~\cite{Goodman:2017}. 
 
Understanding black-box models by approaches like locally fitting a simpler model around the test point~\cite{Ribeiro:2016} or by perturbing the train point to see how the prediction changes~\cite{Simonyan:2013}, \cite{Li:2016}, \cite{Datta:2016} do not satisfactorily indicate where the model came from~\cite{pmlr-v70-koh17a}. To answer this question, the influence function~\cite{Hampel:1974} was introduced; it was a classic technique based on robust statistics through which the learning algorithm can be inspected, and can be traced back to the most influential training data points which impacts the model to predict what it predicts. A simple and efficient methodology was introduced to align and fit the influence function to the machine learning paradigm, which required access to gradients and Hessian-vector products~\cite{pmlr-v70-koh17a}. It was further demonstrated by~\cite{Basu2020} that non-convex and non-differentiable models, which seem to have limited usefulness, successfully provide significant information while approximated by influence function analysis. On linear models, it can be observed that the influence function is useful in -- explaining model predictions, tracking and reducing errors in datasets, debugging models, and even fabricating indistinguishable training set impact\footnote{https://christophm.github.io/interpretable-ml-book/}. The influence function indicates ‘influential’ training data points during model prediction and has a plethora of applications. The authors in~\cite{Han:2020} employed them to explain model predictions and uncover data artifacts. They were used by~\cite{Yang:2020} in order to determine the quality of synthetic training samples within the framework of data augmentation. The authors in~\cite{kobayashi:2020} investigated what would happen if they used gradient-based approaches in conjunction with influence functions to investigate training history and test stimuli simultaneously. One of the drawbacks of influence functions is that it is highly compute intensive. To circumvent this problem FastIf~\cite{guo-etal-2021-fastif}, a collection of simple modifications were proposed to significantly improve the runtime for computing influence functions.

Of late, there have been a rising interest in debugging models using explainability techniques~\cite{Teso:2019,lertvittayakumjorn:2020,guo-etal-2021-fastif, Xu2020OnTI, Nuamah2020ExplainableII,banerjee2021exploring}. In~\cite{Rajani:2020}, the authors suggest utilizing kNN representations to identify training instances responsible for a model's predictions and acquire a corpus-level knowledge of the model's behavior. A recent research~\cite{zylberajch:2021} (HILDIF) has sought to use explainability feedback as input to fine-tune the model for the MNLI dataset. Recently, some comparable tests were carried out using image data, randomly flipping two labels using the influence function~\cite{Hao:2020,teso2021interactive,Wang:2018}. UIDS by~\cite{wang2019less} and RDIA by~\cite{kong:2022resolving}, can both relabel data points based on influence capability using just numeric attributes. To the best of our knowledge, RDIA is the most recent study that addresses the problem of data relabeling followed by a classification task. ~\cite{mozes2023gradientbased} tried to incorporate LLM and utilized influence functions to relabel the predictions. There is a major gap between these works and what we can accomplish with the available textual data. Our work differs from these in that it employs influence functions as a pseudo-annotator and leverages the influential instances as feedback to adjust the gold annotation for a target instance, thereby, improving the overall model performance.
\section{Preliminaries}
\noindent\textbf{Notation}: Let us consider a classification task with input text $t \in \mathcal{T} = \{1,2,...T\}$ and the label $Y=\{y_1,y_2, ..\}$. Each instance $t$ consists of $m$ no. of words, i.e., $t=\{w_1,w_2,...w_m\}$. Let us assume that the feature matrix for the input text $\mathcal{T}$ is $X$. We further denote the training set (texts and their corresponding labels) as $(X_{TR}, Y_{TR})$. In this work, we have multiple validation sets. The validation set will be denoted by $V$. For the test data $X_{TS}$, we have gold labels $Y_{TS}$, and the predicted label will be denoted by $\hat{Y}_{TS}$.\\
\noindent\textbf{Influence function}: Let us choose an instance $(x_i, y_i)$ from $(X_{TR}, Y_{TR})$. Let us have a model $\theta$ and loss functions $\mathcal{L}((x_{i},y_{i}), \theta)$. Given $n$ number of instances in training set $(X_{TR}, Y_{TR})$, our objective is to minimize the loss using $\hat{\theta}= {\arg\min}_{\theta}\frac{1}{n}\sum_{i=1}^{n}\mathcal{L}((x_{i}, y_{i}), \theta)$. Now, the objective attempts to identify the influence of the training data points on the learned parameter $\theta$ and also on the test data $(x_{ts}, y_{ts})\in(X_{TS}, Y_{TS})$.\\
The strength of an influence function is that it attempts to identify the loss locally and tracks the whole model behavior by perturbing or up-weighting it. Let us consider that the loss of a particular training data point is denoted by $\pm{\delta}$. Thus, the influence function for a test data point $(x_{ts}, y_{ts})$ can be represented as follows.
\begin{equation} 
\label{linear}
\footnotesize
IF\{(x_i, y_i), (x_{ts}, y_{ts})\} \cong \frac{d\mathcal{L}((x_{ts}, y_{ts}), \hat{\theta}_{\pm{\delta},(x_i, y_i)})}{d({\pm{\delta}})}
\end{equation}
where $\hat{\theta}_{\pm{\delta},(x_i, y_i)}$ is the model which has been up-weighted or perturbed by $\pm{\delta}$. The updated loss function thus becomes
\begin{equation}
\footnotesize
\label{linear1}
\hat{\theta}= {\arg\min}_{\theta}\frac{1}{n}\sum_{i=1}^{n}\{\mathcal{L}((x_{ts}, y_{ts}), \theta) + ({\pm{\delta}})\mathcal{L}((x_i, y_i), \theta)\}
\end{equation}
\cite{pmlr-v70-koh17a} have shown that to avoid high computation costs, we can compute the influence function using the approximation below.
\begin{equation} 
\footnotesize
\label{linear2}
\begin{split}
& IF\{(x_i, y_i), (x_{ts}, y_{ts})\} \approx -\nabla_\theta \mathcal{L}((x_{ts}, y_{ts}), \hat{\theta})^T {H}_{\hat{\theta}}^{-1}\nabla_\theta\mathcal{L}((x_i,y_i), \hat{\theta})
\end{split}
\end{equation}
where $H_{\hat{\theta}}$ is the Hessian matrix of the model parameters. We are interested in identifying the most negatively influential (helpful) data points by considering the perturbation of a data point that leads to a lower loss in a test data point. Thus, if we denote the most negatively influential (helpful) training data point as $(\hat{x_i}, \hat{y_i})$ then it can be presented as 
\begin{equation} 
\footnotesize
\label{helpful}
(\hat{x_i}, \hat{y_i}) = {\arg\min}_{(x_i, y_i) \in (X_{TR}, Y_{TR})} IF\{(x_i, y_i), (x_{ts}, y_{ts})\}
\end{equation}
According to \cite{guo-etal-2021-fastif} the computation of equation~\ref{helpful} becomes expensive if the dataset size increases. To overcome this issue, instead of searching those data points in the whole set, we search them in a smaller subset considering minimal changes in the nearest neighbors' quality in retrieving influence-worthy data points. Identification of this subset was based on $\emph{l}_2$ distance based on the highly-optimized nearest neighbor search library FAISS \cite{8733051}. So the updated equation becomes 
\begin{equation} 
\footnotesize
\label{updatedhelpful}
(\hat{x_i}, \hat{y_i}) = {\arg\min}_{(x_i, y_i) \in (\hat{X}, \hat{Y})} IF\{(x_i, y_i), (x_{ts}, y_{ts})\}
\end{equation}
where $(\hat{X}, \hat{Y})$ is a subset of $(X, Y)$ computed using FAISS\footnote{https://github.com/facebookresearch/faiss}.\\
\noindent\textbf{Problem definition}: Our objective in this paper is to show that the above influence function formulation proposed in the literature can be used to design a feedback mechanism in a learning model to improve upon the performance in any classification task and, in particular, those that are highly subjective in nature.
Examples of such subjective tasks include hate speech detection, stance classification, sarcasm, and irony detection. Since these tasks are subjective, there might be `impure' instances of data points where there are annotator disagreements. In such cases, the idea is whether one can identify other data points that could potentially influence such impure instances. If this hypothesis is valid, one can determine the influence points for the impure point based on the influence function formulation and use the label information of the influence points as a silver label for the impure instances to improve the overall classification performance. We test this hypothesis by having the silver label as feedback in the model. In the next section, we discuss how we design this feedback mechanism.
\begin{figure*}[t]
\centering
\includegraphics[width=1.0\textwidth]{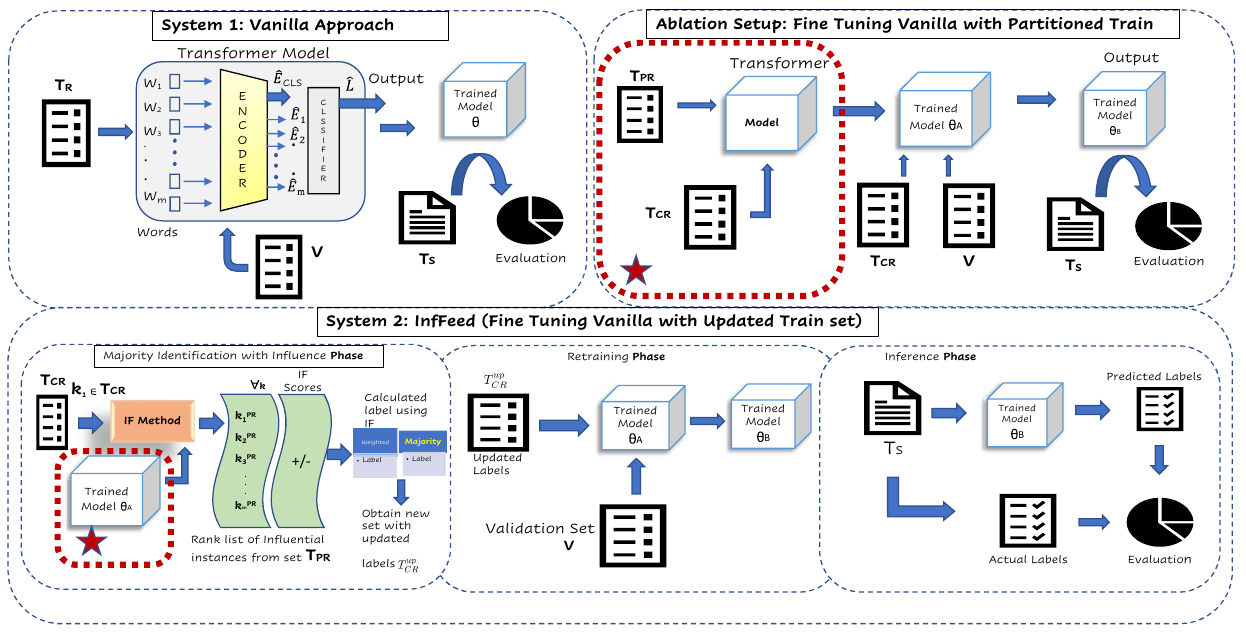}
\caption{Overview of our proposed approach \textsc{InfFeed} along with {\bf System 1} and the vanilla fine-tuning based ablation setup.}
\label{fig2}
\end{figure*}
\section{Methodology}

In this section, we detail the methodology that we adopt to incorporate the influence function as feedback into the classification model. We also discuss the baselines used in this paper. 
\noindent\textbf{Our proposals}: Our proposals include two systems -- \textbf{System 1} and \textbf{System 2}. While \textbf{System 1} is the standard classification model,  \textbf{System 2} is our proposal for incorporating the influence function into the \textbf{System 1}.  
\noindent \textbf{System 1} is the vanilla approach where one usually uses a transformer-based classification model having three divisions of a dataset marked as train $(T_R)$, valid $V$ and test $T_S$. 
We first train a model with $T_R$ and save the model snapshot $\theta$ where the validation loss is minimum and then evaluate the performance using the test data $T_S$. As shown in Figure~\ref{fig2} (\textbf{System 1}) the input text (post/tweet etc.) is split into tokens $\{w_1, w_2, w_3 \cdots w_m\}$ and is passed through a transformer encoder followed by a softmax layer to make the final prediction. 
\if{0}\begin{equation}
        \small
        \label{transencoder}
        \begin{split}
        & \hat{E}_{CLS}, \hat{E}_1, \hat{E}_2 \cdots \\ &\hat{E}_m = TransEnc([E_{CLS}, E_1, E_2 \cdots E_m])
        \end{split}
\end{equation}

\begin{equation}
        \small
        \label{softmax}
        \hat{L} = Softmax(Classifier(\hat{E}_{CLS}))
\end{equation}\fi


\if{0}\noindent \textbf{System 2}: Here we split the training set\footnote{https://docs.cleanlab.ai/v2.0.0/tutorials/pred\_probs\_cross\_val.html} to a smaller subset $T_{CR} \subset T_R$, assume it to be a fine-tuning set, and train a model $(\theta_A)$ only using $T_{PR} = T_R - T_{CR}$. 
Next, we fine-tune $\theta_A$ using $T_{CR}$. 
Subsequently, we use the held-out validation set $V_1$ and save the new model $\theta_B$ where the validation loss is minimum and evaluate the performance with the held-out test set $T_S$. The idea is illustrated in Figure~\ref{fig2} (\textbf{System 2}).\fi \if{0}Mathematically, this can be expressed as follows.
\begin{equation}
        \small
        T_{PR} + T_{CR} \xrightarrow{Transformer\\eqn.~\ref{transencoder},~\ref{softmax}}
        \theta_A
    \end{equation}
    \begin{equation}
        \small
        T_{CR} + V_{1} \xrightarrow{Fine tuning}
        \theta_B
    \end{equation}
    The main purpose of making this particular system is to align the divisions of the dataset and compare apple to apple configurations.\\ \fi
\subsection{Influence function to introduce feedback}
\noindent \textbf{System 2} (\textsc{InfFeed}): We begin by partitioning the training set\footnote{https://docs.cleanlab.ai/v2.0.0/tutorials/pred\_probs\_cross\_val.html}, denoted by $T_R$, into a smaller subset $T_{CR}$, which we designate as a fine-tuning set. Using the remaining part of the training set, $T_{PR} = T_R - T_{CR}$, we then train a model, $\theta_A$. For each instance in $T_{CR}$, we determine the most influential training instances from $T_{PR}$, with $\theta_A$ and the influence function approach outlined in the preceding section. 
We revise the label of each instance in $T_{CR}$ based on the majority/weighted voting of the labels from the top-$K$ influential instances identified earlier, producing an updated set $T_{CR}^{up}$. We proceed to fine-tune $\theta_{A}$ using $T_{CR}^{up}$. Afterwards, we utilize the held-out validation set $V$ to derive the final model, $\theta_B$. Finally, we evaluate $\theta_B$ using the held-out test dataset $T_S$ (Figure~\ref{fig2}, \textbf{System 2}).\\
\noindent\textit{Transformer architectures}: We use the BERT~\cite{devlin} and the DistilBERT~\cite{victorsanh} (a lighter version of BERT) models as transformer architectures throughout this paper. \\
\noindent\textbf{Baselines}: In this paper, we use four state-of-the-art baseline methods taken from the literature - Hao et al.~\cite{Hao:2020}, Rajani et al.~\cite{Rajani:2020}, Wang et al.~\cite{wang2019less}, and Kong et al.~\cite{kong:2022resolving}. As additional baselines, we use two state-of-the-art LLMs GPT-3.5-Turbo\footnote{https://platform.openai.com/docs/models/gpt-3-5} and GPT-4\footnote{https://platform.openai.com/docs/models/gpt-4}, in a zero-shot classification setting. 
\subsection{Influence function to reduce annotation cost}
Imagine a scenario where we have $T_X$ training data points already annotated by human annotators and we wish to enhance the performance of the model by extending the training data with gold annotations of another $T_Y$ points. Rather than having all the $T_Y$ points annotated by the humans, we can use \textsc{InfFeed} to selectively annotate a subset of the $T_Y$ points to reduce the overall annotation cost. To this purpose, we first train the model using $T_X$. Using this trained model we predict the labels for the $T_Y$ points. Thus the $T_Y$ points get silver-annotated. Now we train a fresh model using this silver-annotated $T_Y$ points. For each point in the validation data we get a set of points from $T_Y$ that are most influential using the \textsc{InfFeed} algorithm. Out of these most influential points we concentrate on those that negatively influenced the prediction (had negative influence scores). We ask human annotators to check these cases and, if necessary, re-annotate only these points in $T_Y$. With this revised $T_Y$ we again train the model and find the points negatively influencing the validation data points. Once again these points are re-annotated by humans, if they find it necessary. We repeat this process until in an iteration there are no more negatively influential points.
\if{0}\noindent \textbf{Distant supervision}: We leverage distantly supervised learning to enhance dataset accuracy using influence functions. Initially, we split our training dataset into two equal parts $T_{TR1}$ and $T_{TR2}$, maintaining a balanced representation of each class. We train a model on one half of the dataset~($T_{TR1}$), then use this model to predict labels for the other half~($T_{TR2}$), creating a ``noisy'' set of annotations (we call it distantly annotated). Next, we train a fresh model using this distantly annotated data. After training, influence functions measure the impact of variations in training examples on the validation set~($V$) predictions, aiding in the refinement of noisy labels that significantly affect model performance when compared to the ground-truth annotations. The model undergoes retraining on this refined dataset. We continuously improve its performance using influence functions, label refinement, and retraining, until the performance reaches satisfactory levels. \am{I do not understand this setup at all. Looks like $T_{TR1}=T_{PR}$ and $T_{TR2}=T_{CR}$ thus having no difference with the original setup?}\\ \fi
\if{0}\begin{equation}
        \small
        T_{PR} + T_{CR} \xrightarrow{Transformer\\eqn.~\ref{transencoder},~\ref{softmax}}
        \theta_A
    \end{equation}
    \begin{equation}
        \small
        \begin{split}
        &\forall{l} \in T_{CR}
        \xrightarrow{IF   Method}
        \{i_1^{PR}, i_2^{PR} \cdots i_j^{PR}\}\\ & where   \{j\} \in \{5,10,20,50,100\}
        \end{split}
    \end{equation}
    \begin{equation}
        \small
        T_{CR}^{UP}
        = mode{\{i_1^{PR}, i_2^{PR} \cdots i_j^{PR}\}}
    \end{equation}
    \begin{equation}
        \small
        T_{CR}^{UP} + V_{1} \xrightarrow{Transformer\\ \theta_A}
        \theta_B
    \end{equation}\fi
\if{0}\noindent\textbf{Influence functions can reduce annotation cost}: present a cost-effective solution in machine learning tasks by effectively identifying mislabeled data in training sets. For instance, consider a dataset with 10,000 instances. If manual annotation costs \$1 per instance, the total cost of annotation would be \$10,000. Suppose that the influence function technique identifies 5\% of the data as potentially mislabeled. Reviewing and correcting this subset costs only \$500, a significant reduction compared to the initial cost. Hence, by pinpointing mislabeled instances, influence functions decrease the need for comprehensive manual correction, thereby saving time and significantly reducing annotation costs.\\ \fi
\if{0}\textbf{BERT}     Bidirectional Encoder Representations from Transformers have prior training in the English language, which was obtained from a variety of credible sources.
It has 12 \bm{Use the correct quotations. I have corrected it here.}``attention heads" that are enhanced with a mechanism for self-attention. $BERT_{Base}$ and $BERT_{LARGE}$ are the names given to the two different baseline models that are currently available. \bm{If you are not using bertLarge, don't mention it.}
In order to make our experiment less computationally intensive, we have given some thought to the $BERT_{BASE}$ variable at this point in time.
$BERT_{BASE}$ uses 110M parameters, 12 layers, 768 dimensions, 12 heads, and BERT's base layer count is 12.\\
\textbf{DistilBERT} Although DistilBERT is a lighter variant of BERT, its performance is pretty comparable to that of BERT.
Apparently, the token-type embedding and the pooler are eliminated, and the number of layers is cut down by a factor of two, as stated by ~\cite{victorsanh}.
It's default configuration consists of 768 dimensions over 6 layers and 12 heads.
\\ 
The most important reason for using these two models is that, in general, they are relatively easy to understand and apply, and they have been trained on a wide variety of reliable sources.
One further thing to consider is that our goal is to concentrate mostly on selecting suitable targets from social datasets by using simple models and locating useful, impactful instances that have the potential to influence the game.\bm{I Would need to rephrase this.}\fi
 
\if{0}\noindent\textbf{Hao et al.~\cite{Hao:2020}}: In this work, authors have proposed an automated weakly supervised scheme along with two metric functions for identifying mislabeled data in a binary classification task. The metric functions are cross entropy loss and the influence function. Cross entropy loss is used to calculate the disparity between ground truth and predicted label. The influence function is used to identify the dependence of the model on the training data. Performance is measured after correcting the mislabeled instances. The authors have conducted the experiments on $\sim$10K images from the real-world clinical questions, i.e., mammographic breast density category classification\footnote{http://www.eng.usf.edu/cvprg/Mammography/Database.html} and breast cancer diagnosis. 

\noindent\textbf{Rajani et al.~\cite{Rajani:2020}}: In this work, the authors have proposed a method using $k$-nearest neighbor representations to identify training instances responsible for prediction. Further, they observed that their proposed method is useful for unveiling learned spurious associations, identifying mislabelled instances, and improving model performance. In order to understand the model behavior, kNN was employed over the hidden representation of the model to identify relevant training instances for a test instance. They then identified the confidence interval where kNN performed better than the model. During inference, they either consider the model's prediction or kNN's prediction based on the confidence ranges where each performed better than the other. They have conducted experiments on multiple datasets such as the Stanford Natural Language Inference (SNLI)\footnote{https://nlp.stanford.edu/projects/snli/},  the Adversarial NLI (ANLI)\footnote{https://huggingface.co/datasets/anli} and the Heuristic Analysis for NLI Systems (HANS)\footnote{https://github.com/tommccoy1/hans} datasets.

\noindent\textbf{Wang et al.~\cite{wang2019less}}: In this work, the authors presented a unique Unweighted Influence Data Subsampling (UIDS) approach, and established that the subset-model acquired using the UIDS method can outperform the full-set-model. They separated their whole system into two sections: computing IF and creating probabilistic sampling functions. They created two probabilistic sampling functions, linear sampling (inspired by ~\cite{NEURIPS2018_57c0531e}) and sigmoid sampling. This probabilistic sampling strategy manages the worst-case risk across all distributions that are close to the empirical distribution. They demonstrated their abilities on 14 distinct datasets from the medical, text, social, imaging, Physics, CTR, and life domains.

\noindent\textbf{Kong et al.~\cite{kong:2022resolving}}: In this work, the authors present RDIA, an influence-based relabeling framework for reusing harmful training samples in order to improve model performance. The influence function was used to assess how relabeling a training sample might affect the model's test performance. They conducted their entire experiment on ten distinct datasets (Breast-cancer, Diabetes, News20, Adult, Real-sim, Covtype, Criteo1\%, Avazu, MNIST, CIFAR10)\footnote{https://www.csie.ntu.edu.tw/cjlin/libsvmtools/datasets/} based on a set of numerical features. They employed logistic regression (convex optimization) as the classifier. The average test loss with standard deviation results was used to evaluate performance.

Since UIDS and RDIA models need numerical features as input we obtain pretrained embeddings of all the data points present in our dataset which are then directly fed as input to these models.\fi


\section{Dataset}

The method proposed by us is generic in nature. However, to demonstrate the real effectiveness of the approach, we choose datasets that involve subjective tasks. Our datasets are chosen in a way to cover a wide spectrum of problems and comprise both binary and multiclass scenarios.  
In specific, we focus on four types of subjective tasks -- hate speech detection, stance classification, sarcasm, and irony detection. We evaluate our method on state-of-the-art datasets including -- (a) HateXplain~\cite{hatexplain} and (b) Davidson~\cite{hateoffensive} for hate speech (c) WTWT~\cite{wtwt} and (d)~\cite{mohammad-etal-2016-dataset} for stance classification, (e) isarcasm~\cite{isarcasm} for sarcasm detection, (f)~\cite{van-hee-etal-2018-semeval} for irony detection. The basic statistics for each of these datasets are given in Table~\ref{tab:dataset}. 

\begin{table}
\centering
\resizebox{.67\textwidth}{!}{
\begin{tabular}{|l|c|c|l|} 
\hline
\multicolumn{1}{|c|}{\textbf{Dataset }} & \textbf{Size} & \textbf{\#Labels} & \multicolumn{1}{c|}{\textbf{Name of labels (\#instances)}}                                                                                                                        \\ 
\hline
HateXplain~\cite{hatexplain}                               & 20,148              & 3                      & \begin{tabular}{@{\labelitemi\hspace{\dimexpr\labelsep+0.5\tabcolsep}}l@{}}Hateful (5,935)\\Offensive (5,480)\\Normal (7,814)\end{tabular}                                          \\ 
\hline
HateSpeech~\cite{hateoffensive}                                & 24,802              & 3                      & \begin{tabular}{@{\labelitemi\hspace{\dimexpr\labelsep+0.5\tabcolsep}}l@{}}Hate speech (1,430)\\Offensive (19,190)\\Normal (4,163)\end{tabular}                                      \\
\hline
WT-WT~\cite{wtwt}                                   & 51,284              & 4                      & \begin{tabular}{@{\labelitemi\hspace{\dimexpr\labelsep+0.5\tabcolsep}}l@{}}Support (6,663)\\Refute (4,224)\\Comment (20,864)\\Unrelated (19,533)\end{tabular}                                 \\ 

\hline
Stance~\cite{mohammad-etal-2016-dataset}                                 & 4,163               & 3                      & \begin{tabular}{@{\labelitemi\hspace{\dimexpr\labelsep+0.5\tabcolsep}}l@{}}Favor (1,056)\\Against (2,112)\\Neither (996)\end{tabular}                                             \\ 
\hline
iSarcasm~\cite{isarcasm}                                & 4,484               & 2                      & \begin{tabular}{@{\labelitemi\hspace{\dimexpr\labelsep+0.5\tabcolsep}}l@{}}Sarcastic (777)\\Non-sarcastic (3,707)\end{tabular}                                            \\
\hline
Irony~\cite{van-hee-etal-2018-semeval}                                   & 3,000                & 4                      & \begin{tabular}{@{\labelitemi\hspace{\dimexpr\labelsep+0.5\tabcolsep}}l@{}} Ironic by clash (1,728)\\Situational irony (401)\\Other verbal irony (267)\\Non irony (604)\end{tabular}  \\ 
\hline

\end{tabular}
}
\caption{Dataset details.} 
\label{tab:dataset}
\end{table}

\begin{table*}[ht]
\small
\centering
\scalebox{0.72}{
\begin{tabular}{|c|cccccccccccc|}
\hline
\multirow{3}{*}{\textbf{Setup}}                                              & \multicolumn{2}{c|}{\textbf{HateXplain}}                                    & \multicolumn{2}{c|}{\textbf{WT-WT}}                                          & \multicolumn{2}{c|}{\textbf{IR}}                                           & \multicolumn{2}{c|}{\textbf{ST}}                                           & \multicolumn{2}{c|}{\textbf{iSarcasm}}                                      & \multicolumn{2}{c|}{\textbf{DV}}                        \\ \cline{2-13} 
                                                                             & \multicolumn{12}{c|}{\textbf{Macro F1-score}}                                                                                                                                                                                                                                                                                                                                                                                                                \\ \cline{2-13} 
                                                                             & \multicolumn{12}{c|}{\textbf{Pretrained embedding}}                                                                                                                                                                                                                                                                                                                                                                                                          \\ \hline
\textbf{Wang et al.~\cite{wang2019less} (Lin-UIDS)}                     & \multicolumn{2}{c|}{0.519}                                                      & \multicolumn{2}{c|}{0.490}                                                       & \multicolumn{2}{c|}{0.574}                                                     & \multicolumn{2}{c|}{0.498}                                                     & \multicolumn{2}{c|}{0.502}                                                  & \multicolumn{2}{c|}{0.411}                                  \\ \hline

\textbf{Wang et al.~\cite{wang2019less} (Sig-UIDS)}                     & \multicolumn{2}{c|}{0.562}                                                      & \multicolumn{2}{c|}{0.511}                                                       & \multicolumn{2}{c|}{0.624}                                                     & \multicolumn{2}{c|}{0.523}                                                     & \multicolumn{2}{c|}{0.541}                                                  & \multicolumn{2}{c|}{0.497}                                  \\ \hline

\textbf{Kong et al.~\cite{kong:2022resolving} (RDIA)}                     & \multicolumn{2}{c|}{0.574}                                                  & \multicolumn{2}{c|}{0.536}                                                   & \multicolumn{2}{c|}{0.611}                                                 & \multicolumn{2}{c|}{0.519}                                                 & \multicolumn{2}{c|}{0.546}                                                      & \multicolumn{2}{c|}{0.531}                                  \\ \hline

                                                                             & \multicolumn{1}{c|}{\textbf{BBU}}    & \multicolumn{1}{c|}{\textbf{DB}}     & \multicolumn{1}{c|}{\textbf{BBU}}    & \multicolumn{1}{c|}{\textbf{DB}}      & \multicolumn{1}{c|}{\textbf{BBU}}   & \multicolumn{1}{c|}{\textbf{DB}}     & \multicolumn{1}{c|}{\textbf{BBU}}   & \multicolumn{1}{c|}{\textbf{DB}}     & \multicolumn{1}{c|}{\textbf{BBU}}    & \multicolumn{1}{c|}{\textbf{DB}}     & \multicolumn{1}{c|}{\textbf{BBU}}     & \textbf{DB}     \\ \hline
\textbf{Hao et al.~\cite{Hao:2020}}   & \multicolumn{1}{c|}{0.623}     & \multicolumn{1}{c|}{0.631}           & \multicolumn{1}{c|}{-}               & \multicolumn{1}{c|}{-}                & \multicolumn{1}{c|}{-}              & \multicolumn{1}{c|}{-}               & \multicolumn{1}{c|}{-}              & \multicolumn{1}{c|}{-}               & \multicolumn{1}{c|}{0.598}           & \multicolumn{1}{c|}{0.577}           & \multicolumn{1}{c|}{0.759}            & 0.742           \\ \hline
\textbf{Rajani et al.~\cite{Rajani:2020}} & \multicolumn{1}{c|}{0.611}           & \multicolumn{1}{c|}{0.585}           & \multicolumn{1}{c|}{0.613}     & \multicolumn{1}{c|}{0.603}            & \multicolumn{1}{c|}{\textbf {0.709}}    & \multicolumn{1}{c|}{0.626}           & \multicolumn{1}{c|}{\textbf{0.611}} & \multicolumn{1}{c|}{0.572}           & \multicolumn{1}{c|}{0.515}           & \multicolumn{1}{c|}{0.524}           & \multicolumn{1}{c|}{\underline {0.786}}      & {\underline {0.751}}     \\ \hline
\textbf{{System 1}}                                  & \multicolumn{1}{c|}{0.622}           & \multicolumn{1}{c|}{\underline {0.641}}     & \multicolumn{1}{c|}{0.613}     & \multicolumn{1}{c|}{0.612}      & \multicolumn{1}{c|}{0.683}          & \multicolumn{1}{c|}{0.680}     & \multicolumn{1}{c|}{0.578}          & \multicolumn{1}{c|}{0.588}     & \multicolumn{1}{c|}{0.603}     & \multicolumn{1}{c|}{0.612}     & \multicolumn{1}{c|}{0.765}            & 0.746           \\ \hline

\textbf{{InfFeed (MV)}}    & \multicolumn{1}{c|}{\underline{0.648}}           & \multicolumn{1}{c|}{0.639}           & \multicolumn{1}{c|}{\underline{0.629}}           & \multicolumn{1}{c|}{\underline{0.617}}            & \multicolumn{1}{c|}{\textbf{0.709}} & \multicolumn{1}{c|}{\textbf{0.707*}} & \multicolumn{1}{c|}{\textbf{0.611}} & \multicolumn{1}{c|}{\underline{0.603}} & \multicolumn{1}{c|}{\underline{0.623}}           & \multicolumn{1}{c|}{\underline{0.629}}           & \multicolumn{1}{c|}{0.784}            & 0.749           \\ \hline
\textbf{{InfFeed (WV)}}    & \multicolumn{1}{c|}{\textbf{0.653*}} & \multicolumn{1}{c|}{\textbf{0.657*}} & \multicolumn{1}{c|}{\textbf{0.631}} & \multicolumn{1}{c|}{\textbf{0.622**}} & \multicolumn{1}{c|}{\underline{0.701}}          & \multicolumn{1}{c|}{0.669}           & \multicolumn{1}{c|}{\textbf{0.605*}}          & \multicolumn{1}{c|}{\textbf{0.605}}           & \multicolumn{1}{c|}{\textbf{0.629*}} & \multicolumn{1}{c|}{\textbf{0.635*}} & \multicolumn{1}{c|}{\textbf{0.799**}} & \textbf{0.770*} \\ \hline
& \multicolumn{12}{c|}{\textbf{Large Language Models}}                                                                                                                                                                                                                                                                                                                                                                                                          \\ \hline
\textbf{gpt-3.5-turbo}                     & \multicolumn{2}{c|}{0.638}                                                      & \multicolumn{2}{c|}{\underline{0.629}}                                                       & \multicolumn{2}{c|}{0.682}                                                     & \multicolumn{2}{c|}{0.566}                                                     & \multicolumn{2}{c|}{0.493}                                                  & \multicolumn{2}{c|}{0.735}                                  \\ \hline

\textbf{gpt-4}                     & \multicolumn{2}{c|}{0.644}                                                      & \multicolumn{2}{c|}{\textbf{0.631}}                                                       & \multicolumn{2}{c|}{\underline{0.689}}                                                     & \multicolumn{2}{c|}{0.601}                                                     & \multicolumn{2}{c|}{0.541}                                                  & \multicolumn{2}{c|}{\textbf{0.770}}                                  \\ \hline
\end{tabular}
}
\caption{Macro F1 score for the different models. All bold face entries represent the best performing score and the underlined values represent the best performing baseline. IR: \protect\cite{van-hee-etal-2018-semeval} dataset, ST: \protect\cite{mohammad-etal-2016-dataset} dataset, DV: \protect\cite{hateoffensive} dataset, BBU: BERT-base-uncased, DB: DistilBERT, MV: majority voting, and WV: weighted Voting. *: Statistically significant results with $p$-value $<$0.05, and **: Statistically significant results with $p$-value $<$0.01. Best results are highlighted in bold and second best are underlined.}
\label{tab:systemResults}
\end{table*}

\section{Experimental setup}
We use three different setups in our experiment to observe the importance of increasing data. The setups are as follows -- (i) $S_1$: Here, we randomly sample 2500 instances from the dataset. Then, we split these into four parts : $T_{PR}$ (1000 instances), $T_{CR}$ (800 instances), $V$ (200 instances) and $T_S$ (500 instances). (ii) $S_2$: Here we have 6000 randomly sampled instances and the number of instances in $T_{PR}$, $T_{CR}$, $V$ and $T_S$ are 4200, 800, 500 and 500 respectively. (iii) $S_3$: In this case, the number of randomly sampled instances is 10000. The number of instances in $T_{PR}$, $T_{CR}$, $V$ and $T_S$ are 7500, 1500, 500 and 500 respectively. For each setup, we sample the union of $T_{PR}$, $T_{CR}$, $V$ three times and compute the performance. We keep the test set $T_S$ fixed across all the setups. We take the average of the three macro F1 scores as the final performance. This result is representative, and the trends remain similar for setups with more than 10000 randomly sampled instances. In the case of the datasets which have less number of instances (less than the total instances in $S_2$ but more than $S_1$), we oversample the instances in training data ($T_{PR}$) using random selection with repetition. 

For the baselines~\cite{Hao:2020,Rajani:2020,wang2019less,kong:2022resolving} also, we have three such setups; however, during training, we merge $T_{PR}$ and $T_{CR}$ to form a single training set. We let the validation ($V$) and test ($T_S$) sets remain the same. For the LLM baselines we query the models with each entry from the test set $T_S$ and record the classification labels in each case.\\
\noindent\textbf{Model setup}: For \textbf{System 1} and \textbf{System 2} (i.e., \textsc{InfFeed}), we have used two models -- BERT-base and DistilBERT. During the fine-tuning, we freeze the first nine layers based on the findings in~\cite{Lee2019} to limit the amount of computation. This leaves us with approximately 14.7M trainable parameters. In the case of DistilBERT, we freeze the first 4 layers to bring down the overall computation cost. For both models, we consider a maximum of 350 tokens. After parameter tuning, the learning rate is set at $2e-5$, the number of epochs at 12, and the batch size at 64. Further, for \textsc{InfFeed},  the weight decay is set to 0.005, the $k$ in kNN to 100, and the Hessian approximation value to 800. 

For~\cite{Hao:2020}, everything else remaining same as \textbf{System 1}, the learning rate has been set to $5e-5$. In the case of this baseline, we treat the hate speech datasets as a two-class classification scenario whereby we merge the `hateful' and the `offensive' classes into a single `abusive' class. Now, during classification, we randomly select 10\% of the instances from the entire dataset along with their original labels; we then flip the label for each instance to `abusive' if the original label is `normal' and vice versa. We did the same for the sarcasm dataset. For~\cite{Rajani:2020}, the learning rate and the $k$ in kNN have been set to $5e-5$ and 16, respectively, while everything else remains the same as \textbf{System 1}. 

For the baselines UIDS~\cite{wang2019less} and RDIA~\cite{kong:2022resolving} we use the Newton-CG algorithm~\cite{Marten:2010} to calculate Influence Functions as mentioned in the paper. For the logistic regression model mentioned in RDIA, we select the regularization term $C = 0.1$.\\
\noindent\textbf{System setup}: We run all of the models described in this study on a Windows-based system equipped with 64 gigabytes of RAM, two 24 gigabytes RTX 3090 GPU connected through SLI, and a Ryzen 9 with a fifth generation, twelve-core CPU.

\subsection{Description of the baselines}
\noindent\textbf{Hao et al.~\cite{Hao:2020}}: In this work, authors have proposed an automated weakly supervised scheme along with two metric functions for identifying mislabeled data in a binary classification task. The metric functions are cross entropy loss and the influence function. Cross entropy loss is used to calculate the disparity between ground truth and predicted label. The influence function is used to identify the dependence of the model on the training data. Performance is measured after correcting the mislabeled instances. The authors have conducted the experiments on $\sim$10K images from the real-world clinical questions, i.e., mammographic breast density category classification\footnote{http://www.eng.usf.edu/cvprg/Mammography/
Database.html} and breast cancer diagnosis. 

\noindent\textbf{Rajani et al.~\cite{Rajani:2020}}: In this work, the authors have proposed a method using $k$-nearest neighbor representations to identify training instances responsible for prediction. Further, they observed that their proposed method is useful for unveiling learned spurious associations, identifying mislabelled instances, and improving model performance. In order to understand the model behavior, kNN was employed over the hidden representation of the model to identify relevant training instances for a test instance. They then identified the confidence interval where kNN performed better than the model. During inference, they either consider the model's prediction or kNN's prediction based on the confidence ranges where each performed better than the other. They have conducted experiments on multiple datasets such as the Stanford Natural Language Inference (SNLI)\footnote{https://nlp.stanford.edu/projects/snli/},  the Adversarial NLI (ANLI)\footnote{https://huggingface.co/datasets/anli} and the Heuristic Analysis for NLI Systems (HANS)\footnote{https://github.com/tommccoy1/hans} datasets.

\noindent\textbf{Wang et al.~\cite{wang2019less}}: In this work, the authors presented a unique Unweighted Influence Data Subsampling (UIDS) approach, and established that the subset-model acquired using the UIDS method can outperform the full-set-model. They separated their whole system into two sections: computing IF and creating probabilistic sampling functions. They created two probabilistic sampling functions, linear sampling (inspired by ~\cite{NEURIPS2018_57c0531e}) and sigmoid sampling. This probabilistic sampling strategy manages the worst-case risk across all distributions that are close to the empirical distribution. They demonstrated their abilities on 14 distinct datasets from the medical, text, social, imaging, Physics, CTR, and life domains.

\noindent\textbf{Kong et al.~\cite{kong:2022resolving}}: In this work, the authors present RDIA, an influence-based relabeling framework for reusing harmful training samples in order to improve model performance. The influence function was used to assess how relabeling a training sample might affect the model's test performance. They conducted their entire experiment on ten distinct datasets (Breast-cancer, Diabetes, News20, Adult, Real-sim, Covtype, Criteo1\%, Avazu, MNIST, CIFAR10)\footnote{https://www.csie.ntu.edu.tw/cjlin/libsvmtools/datasets/} based on a set of numerical features. They employed logistic regression (convex optimization) as the classifier. The average test loss with standard deviation results was used to evaluate performance.

Since UIDS and RDIA models need numerical features as input we obtain pretrained embeddings of all the data points present in our dataset which are then directly fed as input to these models.
\section{Influence function as a feedback}
\begin{figure}[t]
\centering
\includegraphics[width=10cm,height=7cm]{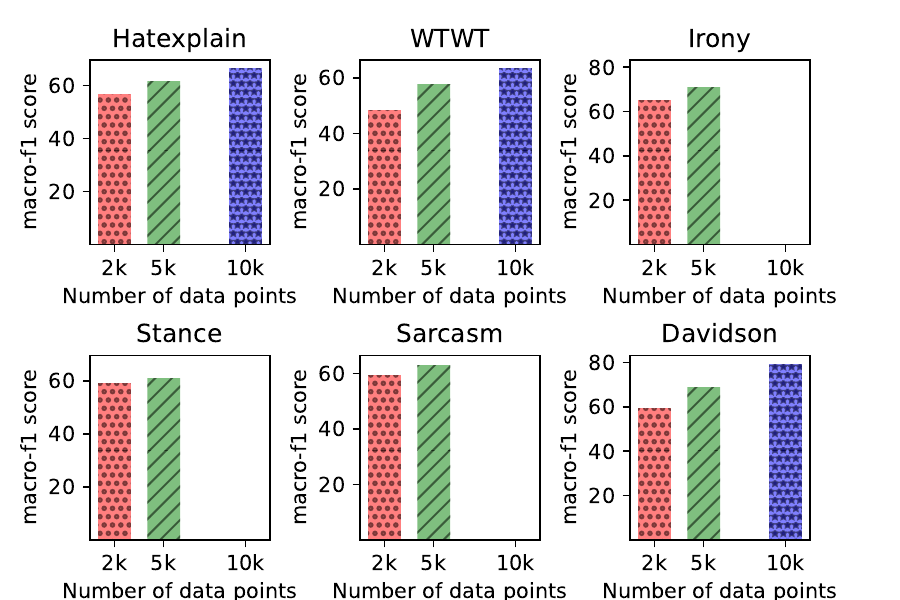}
\caption{Demonstration of how the macro F1-score improves with increasing data. Stance: \protect\cite{mohammad-etal-2016-dataset} dataset, Irony: \protect\cite{van-hee-etal-2018-semeval} dataset, and Davidson:\protect\cite{hateoffensive} dataset.}
\label{performancewithsize}
\end{figure}

In Table~\ref{tab:systemResults}, we summarize our main results. As our dataset does not have numerical features, we represent the data points using BERT based pretrained embeddings that are fed to UIDS and RDIA as inputs.
 The \textbf{BBU} and \textbf{DB} columns show the results using BERT-base-uncased and DistilBERT as the transformer architectures, respectively. 
All the results are averaged over the three setups $S_1$, $S_2$ and $S_3$. We observe that \textsc{InfFeed} (majority/weighted voting) always outperforms the most competing baselines except for the~\cite{mohammad-etal-2016-dataset} dataset, where it is the same as the baseline. In all cases where our models win, the results are statistically significant. In general, \textsc{InfFeed} weighted voting is slightly better than majority voting. Further, for both \textsc{InfFeed} models, the DistilBERT architecture performs better than BERT-base-uncased in most cases. For the baselines~\cite{Hao:2020} and~\cite{Rajani:2020}, the trends are reversed; BERT-base-uncased generally works better than DistilBERT here.\\
Our models also outperform the LLM based baselines. The largest performance margin is for the iSarcasm dataset with GPT-4 reporting a macro F1 score of 0.541 compared to \textsc{InfFeed} (WV) at 0.635.\\ 
\noindent\textbf{Effect of varying data size}: Here we report the performance of the best performing model, \textsc{InfFeed} (majority voting) separately for the three setups -- $S_1$, $S_2$ and $S_3$. Figure~\ref{performancewithsize}, shows how the performance of the model improves as we increase the dataset size. For some datasets, e.g.,~\cite{hateoffensive} and~\cite{wtwt}, one observes a gain close to 20\% as one sweeps from setup $S_1$ to $S_3$.\\
\noindent\textit{Remark}: According to the study by~\cite{pmlr-v70-koh17a}, with $N$ training data points and $P$ parameters, the Hessian matrix computation requires $O(NP^2 + P^3)$ operations, which is unacceptably expensive for massive datasets/models. This is the primary reason for the popularity of the FastIf~\cite{guo-etal-2021-fastif} algorithm which is also what we have used here.\\
\noindent\textbf{Ablation studies}: In order to understand the effectiveness of the influence function as a `pseudo-expert' annotator, we perform two ablation experiments. These are -- (a) random flipping and (b) vanilla fine-tuning.\\
\noindent\textit{Random flipping}: This system uses the same parameters as mentioned in {\bf System 1}. However, here we randomly flip the labels of some of the training instances (around 5\%, which is similar in tune to the number of instances updated on average by \textsc{InfFeed}).\\
\noindent\textit{Vanilla fine-tuning}: As in \textbf{System 2}, here also we obtain a model $\theta_A$ by training it on $T_{CR}$. Now rather than computing influence functions, we fine-tune $\theta_A$ using $T_{CR}$.\\ 
Subsequently, we use the held-out validation set $V$ and save the new model $\theta_B$ where the validation loss is minimum and evaluate the performance with the held-out test set $T_S$.\\
 The results from the two ablations are reported in Table~\ref{tab:ablation}. For {\em random flipping}, in case of the hate speech datasets, there is an average performance drop of almost 20\%. For the stance detection datasets, we can see an average 16\% drop, while for the irony and sarcasm datasets, the average drops are nearly 13\% and 18\%, respectively. In {\em vanilla fine-tuning} for all the datasets we see an average drop in the range of 2\% -- 2.5\%. Clearly, both the approaches perform worse than \textsc{InfFeed} showing the effectiveness of the influence functions.\\
\noindent\textbf{Example instances}: In Table~\ref{tab:changeableLabel} we show some examples where the incorrect original label gets updated to the correct label based on the votes from the influential instances. This is one of the basic reasons for the better performance of our models.

\begin{table*}[ht]
\small
\centering
\scalebox{0.83}{
\begin{tabular}{|c|llllllllllll|}
\hline
\multicolumn{1}{|c|}{\multirow{3}{*}{\textbf{Setup}}} & \multicolumn{2}{c|}{\textbf{HateXplain}}                                                                                    & \multicolumn{2}{c|}{\textbf{WT-WT}}                                                & \multicolumn{2}{c|}{\textbf{IR}}                                                & \multicolumn{2}{c|}{\textbf{ST}}                                           & \multicolumn{2}{c|}{\textbf{iSarcasm}}                                         & \multicolumn{2}{c|}{\textbf{DV}}                    \\ \cline{2-13} 
\multicolumn{1}{|c|}{}                       & \multicolumn{12}{c|}{\textbf{Macro F1-score}}                                                                                                                                                                                                                                                                                                                                                                                                                                                            \\ \cline{2-13} 
\multicolumn{1}{|c|}{}                       & \multicolumn{1}{l|}{\textbf{BBU}}            & \multicolumn{1}{l|}{\textbf{DB}}  & \multicolumn{1}{l|}{\textbf{BBU}}            & \multicolumn{1}{l|}{\textbf{DB}}     & \multicolumn{1}{l|}{\textbf{BBU}}            & \multicolumn{1}{l|}{\textbf{DB}}     & \multicolumn{1}{l|}{\textbf{BBU}}   & \multicolumn{1}{l|}{\textbf{DB}} & \multicolumn{1}{l|}{\textbf{BBU}}   & \multicolumn{1}{l|}{\textbf{DB}} & \multicolumn{1}{l|}{\textbf{BBU}}   & \textbf{DB} \\ \hline
\textbf{Random flipping}                         & \multicolumn{1}{l|}{0.543}          & \multicolumn{1}{l|}{0.510}                       & \multicolumn{1}{l|}{0.497}          & \multicolumn{1}{l|}{0.432}          & \multicolumn{1}{l|}{0.590}          & \multicolumn{1}{l|}{0.553}          & \multicolumn{1}{l|}{0.419}          & \multicolumn{1}{l|}{0.397}               & \multicolumn{1}{l|}{0.442}          & \multicolumn{1}{l|}{0.437}               & \multicolumn{1}{l|}{0.543}          & 0.523               \\ \hline
\textbf{Vanilla fine-tuning}                    & \multicolumn{1}{l|}{0.637}          & \multicolumn{1}{l|}{0.656}                      & \multicolumn{1}{l|}{0.612}          & \multicolumn{1}{l|}{0.616}          & \multicolumn{1}{l|}{0.691}          & \multicolumn{1}{l|}{0.694}          & \multicolumn{1}{l|}{0.589}          & \multicolumn{1}{l|}{0.599}               & \multicolumn{1}{l|}{0.588}          & \multicolumn{1}{l|}{0.621}               & \multicolumn{1}{l|}{0.770}          & 0.749               \\ \hline
\end{tabular}
}
\caption{Ablation studies. IR:\protect\cite{van-hee-etal-2018-semeval} dataset, ST:\protect\cite{mohammad-etal-2016-dataset} dataset, DV:\protect\cite{hateoffensive} dataset, BBU: BERT-base-uncased, DB: DistilBERT.}
\label{tab:ablation}
\end{table*}

\if{0}\begin{table*}[ht]
\centering
\scalebox{0.59}{
\begin{tabular}{|c|c|c|} 
\hline
\textbf{Dataset}    & \textbf{Target instance}                                                                                                                                                                                                                                                                                                                                                      & \textbf{Three most influential instances}                                                                                                                                                                                                                                                                                                                                                                                                                                                                                          \\ 
\hline
\textbf{HateXplain} & \begin{tabular}[c]{@{}l@{}}\textless{}user\textgreater{} \textless{}user\textgreater{} \textless{}user\textgreater{} \textless{}percent\textgreater{} of ww voted trump \hl{blame all white women} \textless{}number\textgreater{} million registered \\dems could not \hl{get off their asses} to vote blame lazy democrats lol stop whining about votes that did not matter\end{tabular} & \begin{tabular}[c]{@{}l@{}}1. \textless{}user\textgreater{} \textless{}user\textgreater{} genuine question do you \hl{hate white people} as much\\~as it seems you do on twitter\\2. \textless{}user\textgreater{} \textless{}user\textgreater{} \textless{}user\textgreater{} tour guide and here you will see true caucasians\\~just \hl{being white}\\3. \textless{}user\textgreater{} now it a humanitarian crisis because he wants to declare an\\~emergency he doesnt see refugees as \hl{humans resign trump}\end{tabular}  \\ 
\hline
\textbf{Stance}     & Our cause is making definite marks. Never stop                                                                                                                                                                                                                                                                                                                                   & \begin{tabular}[c]{@{}l@{}}1. "Lets all stop saying people don't ""believe"" in Climate Change. \\Its not a matter of faith. They dont UNDERSTAND\\2. There has never been a better time to get the heck out of California.\\3. user Also im assuming you claim to be one...yet you just called me an\\~idiot and you dont even know me. Nice.\end{tabular}                                                                                                                                                           \\ 
\hline
\textbf{Irony}      & I just love when you test my patience!                                                                                                                                                                                                                                                                                                                                           & \begin{tabular}[c]{@{}l@{}}1. user Im hurt but I still love you\\2. Oh sprints I love you\\3. You are such a great person\end{tabular}                                                                                                                                                                                                                                                                                                                                                                              \\
\hline
\end{tabular}
}
\caption{Sample influential instances for a given instance. Stance:~\cite{mohammad-etal-2016-dataset} dataset, Irony:~\cite{van-hee-etal-2018-semeval} dataset. The similarity in context is shown by the portions highlighted in yellow.}
\label{tab:sampleinfluences}
\end{table*}\fi

\begin{table}
\centering
\scalebox{.85}{
\begin{tabular}{|l|l|l|l} 
\cline{1-3}
\textbf{Instances}                                                    & \textbf{Original label} & \textbf{Influence label} &   \\ 
\cline{1-3}
\textless{}user\textgreater{} I love\~ u more fa**ot                                  & Normal                  & Offensive                &   \\ 
\cline{1-3}
\begin{tabular}[c]{@{}l@{}}moslem ban and libtard ban \\but not gun ban\end{tabular} & Offensive               & Normal                   &   \\ 
\cline{1-3}
It was her choice Mark                                                               & Favor                   & None                     &   \\
\cline{1-3}
\begin{tabular}[c]{@{}l@{}}\textless{}user\textgreater{} so i got called a nappy \\headed ni**er jew\end{tabular} & Normal & Offensive    &   \\
\cline{1-3}
make nazi scum lose their jobs   &Normal  &Offensive & \\
\cline{1-3}
\begin{tabular}[c]{@{}l@{}}good these mcu fa**ots gonna \\get schooled on what\end{tabular}  &Normal  &Hatespeech  &  \\
\cline{1-3}
\begin{tabular}[c]{@{}l@{}}that shit doesnt even make sense \\so yes for wetback lmao\end{tabular}  &Hatespeech  &Normal  \\
\cline{1-3}
\begin{tabular}[c]{@{}l@{}}the white bitch amber guyger \\has been locked away\end{tabular} &Normal &Offensive & \\
\cline{1-3}
\end{tabular}
}
\caption{Samples re-labelled.}
\label{tab:changeableLabel}
\end{table}

\section{\textsc{InfFeed} to reduce annotation cost}

\noindent\textbf{Experimental setup}: For all the datasets, we use half of $T_R$ as $T_X$ and the other half as $T_Y$. The validation and the test data are the same as earlier, i.e., $V$ and $T_S$.\\
\noindent\textbf{Results}: We compare the performance of the \textbf{BBU} model trained on $T_Y$ with all gold annotations ($T_Y^\textrm{GOLD}$), the raw silver annotations of $T_Y$ using the model trained with $T_X$ ($T_Y^\textrm{SILVER}$), and the selectively gold annotated $T_Y$ ($T_Y^\textrm{\textsc{InfFeed}}$) using the \textsc{InfFeed} algorithm repeatedly. The results are shown in Table~\ref{table:correction_comparison}. We observe that the results obtained using $T_Y^\textrm{\textsc{InfFeed}}$ are very close to $T_Y^\textrm{GOLD}$ and the results from $T_Y^\textrm{SILVER}$ are inferior to both of these (except for the iSarcasm dataset). For each dataset, the number of data points in $T_Y$ that had to be re-annotated in total are exceptionally low compared to size of $T_Y^\textrm{GOLD}$.

\begin{table}[h]
\centering
\scriptsize
\begin{tabular}{|c|c|c|c|c|}
\hline
\centering
\textbf{Dataset} & $T_Y^\textrm{SILVER}$ & $T_Y^\textrm{\textsc{InfFeed}}$ &  $T_Y^\textrm{GOLD}$ & \#re-annotated\\ \hline
\textbf{HateXplain} &  61 & 65 & 67 & 17\\ \hline
\textbf{WT-WT} & 57 & 60 & 61.5 & 9\\ \hline
\textbf{IR} & 66 & 67 & 70 & 11\\ \hline
\textbf{ST} & 46 & 48 & 55 & 17\\ \hline
\textbf{iSarcasm} & 59 & 59 & 61 & 7\\ \hline
\textbf{DV} & 74 & 75 & 77 & 21\\ \hline
\end{tabular}
\caption{Comparison of model performance in terms of \% accuracy.}
\label{table:correction_comparison}
\end{table}

\if{0}\begin{equation}
        \small
        T_{PR} + T_{CR} \xrightarrow{Transformer\\eqn.~\ref{transencoder},~\ref{softmax}}
        \theta_A
    \end{equation}
    \begin{equation}
        \small
        \begin{split}
        &\forall{l} \in T_{CR}
        \xrightarrow{IF   Method}
        \{i_1^{PR}, i_2^{PR} \cdots i_j^{PR}\}\\ & where   \{j\} \in \{5,10,20,50,100\}
        \end{split}
    \end{equation}
    \begin{equation}
        \small
        T_{CR}^{UP}
        = mode{\{i_1^{PR}, i_2^{PR} \cdots i_j^{PR}\}}
    \end{equation}
    \begin{equation}
        \small
        T_{CR}^{UP} + V_{1} \xrightarrow{Transformer\\ \theta_A}
        \theta_B
    \end{equation}\fi

\section{Error analysis}

In this section, we analyze the incorrect predictions from \textsc{InfFeed} and group them into the following types. 
 Here we identify the frequent mistakes made by the model and place them into different categories -- \textbf{(i)} Presence of slur words may not always correspond to the text being hateful/offensive. Thus, since the model observes such examples in both classes during training it is sometimes confused during prediction. For instance, in the HateXplain dataset, the text ``kevin gates make music for dykes and chubby ni**a'' is wrongly marked as hate speech by the model owing to the presence of the word ``ni**a'', \textbf{(ii)} When the major text feature appears only once in a sentence, that feature is given higher importance during annotation. For example, in the~\cite{van-hee-etal-2018-semeval} dataset, the statement ``Most of us didn’t focus in the \#ADHD lecture'' is labelled as situational irony while the correct class should be other verbal irony. The influential instances also support this same label even though they do not have any contextual overlap, \textbf{(iii)} If there are several targets in the dataset, one individual target instance may be taking influence from another target cluster and producing the incorrect collection of influential instances. For example, in the~\cite{mohammad-etal-2016-dataset} dataset, ``Jeb Bush is the only sane candidate in this republican lineup'' has `Donald Trump' as the target but is influenced by ``Let\'s hope the VOTERS remember" which has `Hilary Clinton' as the target. This may result in possible incorrect predictions, \textbf{(iv)} If a post's contextual information and some label-related input are limited, influential instances are often incapable of capturing the latent semantics of the post. They rather pick up cues from the surface structure. As an example, ``Thank \@user for being so entertaining at the Edinburgh signings! You did not disappoint! I made my flight so will have plenty time to read \@user'' is influenced by instances where the words `entertainment' or `flight' appear in an instance, and \textbf{(v)} If the gold annotations are incorrect, influence functions may relate to the wrong instance, resulting in incorrect voting. For example, in the HateXplain dataset, ``bare naked islam brought to you by bare naked ladies'' is annotated as `hate speech' while it should be actually annotated as `offensive'. If this instance is an influencer for a particular target instance then the silver label of the target instance might get incorrectly updated resulting in possible incorrect prediction. 

\section{Conclusion}
We present \textsc{InfFeed}, which, by leveraging influence as feedback, attempts to simulate a pseudo-expert annotator by updating the label of a target instance. This simple approach results in significantly better performance as compared to the state-of-the-art baselines for a series of classification tasks that are subjective in nature. In the dataset extension setting, we observe that even by manually annotating $\sim\frac{1}{1000}^\textrm{th}$ of the full dataset that need to be extended we obtain comparable performance with the scenario where all the dataset to be extended is gold-annotated. 
In the future, we would like to investigate if this scheme can be effectively used to replace the need for an expert annotator in a real-world deployment scenario through faster computation.

\section{Ethics statement}
In our research, we responsibly use social subjective data, originally published in another study and used with appropriate permissions. Acknowledging the sensitive nature of this data, we have undertaken diligent steps to maintain ethical standards. Specifically, we employed expert annotators to revisit and correct any potential misannotations, enhancing the reliability of our data. This process reinforces our commitment to upholding stringent ethical guidelines in our research.

\bibliographystyle{plainnat}
\bibliography{references}







\end{document}